\documentclass{article}

  \usepackage[preprint]{neurips_2026}

\usepackage[utf8]{inputenc} 
\usepackage[T1]{fontenc}    
\usepackage{hyperref}       
\usepackage{url}            
\usepackage{booktabs}       
\usepackage{amsmath,amsfonts,amsthm} 
\usepackage{bm}
\usepackage{nicefrac}       
\usepackage{microtype}      
\usepackage{xcolor}         
\usepackage{graphicx}
\newcommand{\bs}{\boldsymbol}

\newcommand{\pf}[2]{\frac{\partial #1}{\partial #2}}

\newcommand{\pft}[2]{\frac{\partial^2 #1}{\partial #2^2}}

\newcommand{\df}[2]{\frac{\mathrm{d} #1}{\mathrm{d}  #2}}

\newcommand{\dft}[2]{\frac{\mathrm{d} ^2 #1}{\mathrm{d}  #2^2}}
\newcommand{\md}{\mathrm{d}}

\newcommand{\uk}{\hat{u}_{\bs{k}} }
\newcommand{\mk}{|\bs{k}| }

\title{What Should a Large Language Model See? Physical Invariants as a Data Representation for PDE Discovery}

\author{%
  Fan Yang\thanks{fy2@caltech.edu} \qquad Matt Thomson \\
  Division of Biology and Biological Engineering\\
  California Institute of Technology\\
  Pasadena, CA 91125 \\
}

\begin{document}

\maketitle

\begin{abstract}
Understanding how molecular interactions govern macroscopic behaviour is a central challenge in  molecular sciences.  However, conventional theory building cannot keep pace with the vast datasets modern experimentation  routinely produces. Large language models offer a promising route to automating theory construction, but a spatiotemporal field cannot be directly placed in a prompt. Existing models generally learn about the data only through a score measuring how well each proposal fits it. Here we introduce data interpretation, a stage that measures the field into the quantities a theorist would consult and supplies them to the model as a direct input. On a benchmark of simulated fields, interpretation nearly triples the accuracy of recovered equations relative to showing the raw data, at negligible computational cost and without any training. By allowing a language model to read field data as a theorist does, data interpretation offers a practical route to automated field theory construction that can coevolve with experimentation.
\end{abstract}

\section{Introduction}

 Today, deriving a theoretical model for a complex chemical system typically requires months to years of specialized expertise, while experimentalists can routinely generate thousands of distinct systems by varying and recombining molecular components. This gap arises from the  bottleneck  of human bandwidth. Conventional modeling approaches fall into two categories: bottom-up methods that coarse-grain microscopic interactions through kinetic theory into continuum field equations, and top-down methods that enumerate all symmetry-permitted terms without reference to molecular details. Both approaches are slow, expertise-dependent, and unable to match the volume and complexity of data that modern experimentation now routinely produces. AI, with its capacity for pattern recognition, feature extraction, and learning complex mappings across scientific domains, offers a compelling path toward automating this theory-building process — at the scale and speed that molecular sciences now critically demand.

In this  paper, we focus on automating the top-down approach: recovering partial differential equation (PDE)-based field theories from data.  This is an important task in scientific machine learning, and the dominant formulation has been sparse regression. SINDy \cite{brunton16} and PDE-FIND \cite{rudy17} assemble a library of candidate terms into a design matrix, apply numerical regression to find the best-fit candidates, and use a sparsity penalty to limit the length of PDEs. However, this approach is limited by the library of derivative candidates, as the PDE terms must lie within the library written down in advance.

\begin{figure}[htbp]
	\centering
	\includegraphics[width=0.8\linewidth,angle=0]{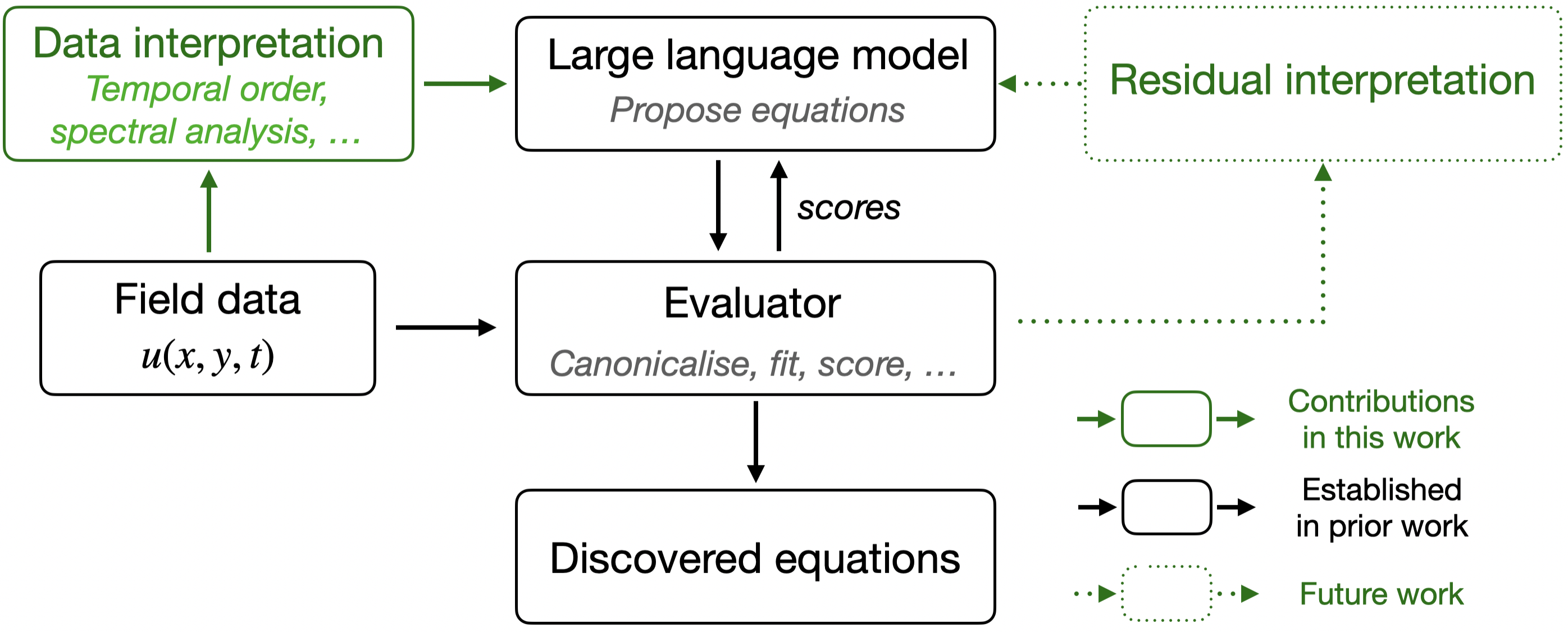}
	\caption{\textbf{Architecture of symbolic regression to discover PDE-based field theory from data.} Established methods pair an LLM with an evaluator: the LLM proposes a candidate equation, and the evaluator fits its coefficients and returns a score. Because the field data typically cannot be placed in a prompt, that score is what informs the LLM about the data. We contribute two  interpretation stages, shown in green. Data interpretation measures the observed field into physical quantities a theorist would compute. In our ongoing work, we apply  residual interpretation to analyze  what a proposal failed to explain. No training is required in this pipeline.}
	\label{fig1}
\end{figure}

Symbolic regression removes the library constraint at the cost of a combinatorial search. Recent progress shows that pretrained large language models (LLMs), equipped with strong  scientific reasoning capabilities, have great potential as the proposal mechanism for symbolic regression.  FunSearch \cite{paredes24} established the scheme of an LLM proposing candidates, a numerical evaluator scoring them, and an evolutionary algorithm driving iterative refinement. LLM-SR \cite{shojaee25} applied this LLM-guided evolutionary approach to scientific equation discovery. Since then, many methods have been proposed to refine either the search or the feedback signal: in addition to evolving the equations, LASR \cite{grayeli24} also evolves a library of hypotheses that can  guide the equation discovery; DrSR \cite{wang25} adds a data analysis LLM to learn data insights, and reflective feedback to store reusable strategies; IGSR \cite{saveliev26} expands the evaluation metrics by scoring per-term influence inside a tree search; LLM-ACES \cite{abhyankar26} incorporated active data acquisition to better evaluate and evolve candidate hypotheses;  MARICL \cite{rezaei26} applies  agents to read high-error examples from a base model and propose improvements. These methods share a core architecture, shown in black in Fig. \ref{fig1}. A pre-trained LLM proposes an equation, a numerical evaluator evaluates it, and the resulting score returns to the LLM to generate new candidates in a loop. Within this architecture, the LLM proposes the symbolic equation structures, while numerical regression fits only the coefficients of the proposed PDE structure. Such a division of labor eliminates the need for a predefined library for PDE term candidates. Effort in the field has mainly concentrated on the evaluator and the search, through improved differentiation schemes, residual formulations, credit assignment, and exploration strategies. 

We focus on a distinct aspect that has received less attention: how the data is represented to the LLM. We are motivated by how human theorists build theories from field data --- by identifying salient structures  in the data. For example, an experienced theorist knows that a shock wave indicates a strong convection effect, and that a Turing pattern signals a diffusion-reaction system. Neither inference requires numerical values of the field, only an interpretation of its pattern.  We therefore propose supplying such an interpretation stage to the model as an input in its own right, rather than solely relying on the evaluator.

We introduce a stage called data interpretation, whose output is supplied directly to the LLM, as shown in green in Fig. \ref{fig1}. Feeding the raw field data  to the LLM is prohibitively expensive, so most existing research uses it only within the evaluator. Data interpretation instead lets the LLM read the data as a theorist does. A  set of measurements is computed and supplied as evidence, chosen to be the quantities a theorist would consult  when encountering an unfamiliar field.  These diagnostics are cheap and compact:  they take a small fraction of the time for a single model call, and occupy fewer tokens than the raw data they replace. They are also interpretable, so what the LLM was told can be audited by both human and AI.  The present work implements a deliberately small set of diagnostics, which are sufficient to test the efficacy of the proposal rather than to exhaust it. In our future work, we will extend the interpretation module with more comprehensive  diagnostics.



\section{Methodology}

The pipeline of PDE discovery consists of four stages in a loop: ``Data interpretation $\to$ LLM proposing an equation $\to$ Parsing and canonicalisation $\to$ Evaluation", as shown in Fig. \ref{fig1}. All experiments in this paper use QwQ-32B. The model is frozen throughout.

\subsection{Data interpretation} \label{section::dataint}

We apply spectral analysis as the main component of data interpretation. We consider a scalar field  $u(\bs{x}, t)$ on a periodic domain, governed by a  constant-coefficient PDE of first or second order in time.  The Fourier transform
\begin{equation}
	\uk(t) = \int u(\bs{x},t) e^{-i\bs{k}\cdot \bs{x}}\md \bs{x}
\end{equation}
can decouple the field data $u(\bs{x}, t)$ into a series of spatial modes $\uk(t)$ at wavevector $\bs{k}$. For a linear PDE,   each mode  $\uk(t)$  evolves independently, obeying an ordinary differential equation whose parameters are the coefficients we seek. The LLM can therefore infer the coefficients directly from how each mode grows, decays, or oscillates. On the contrary, if energy transfers between modes, the decoupling will break down, which indicates a nonlinear PDE. 

The interpretation addresses four questions, each with corresponding measurements, as listed below.
 
 \textbf{Q1: First or second order in time? }For a PDE of first order in time, e.g., $\frac{\partial u}{\partial t}=\nu \nabla^2 u + \sigma u$, where $\nu$ and $\sigma$ are coefficients, each Fourier mode satisfies
\begin{equation}
	\frac{\mathrm{d} \hat{u}_{\bs{k}} }{\mathrm{d}t}=(\sigma -\nu |\bs{k}|^2)  \hat{u}_{\bs{k}} , \qquad \mbox{so} \quad \uk(t) = \uk(0)e^{(\sigma -\nu |\bs{k}|^2)t}, \label{uk::firstorder}
\end{equation}
 Therefore, the amplitude $|\uk(t)|$ is monotonic in time. 
 
 For a PDE of second order in time, e.g., $\pft{u}{t}=c^2\nabla^2 u + \sigma u + \gamma \pf{u}{t}$, each mode is  instead a damped oscillator
 \begin{equation}
 	\dft{\uk}{t}-\gamma \df{\uk}{t} + \left(c^2\mk^2-\sigma\right)\uk=0, \label{uk::2ndorder}
 \end{equation}
 and  the amplitude $|\uk(t)|$ is oscillatory  in time. 

We measure the time evolution of $|\uk(t)|$  to infer the temporal order of the PDE. However, this test is not exhaustive. For example, an overdamped second-order equation, with $\gamma^2\geq 4(c^2\mk^2-\sigma)$ decays monotonically and is indistinguishable from a first-order one. This shows that data interpretation constrains the space of plausible structures rather than determining one. 

\textbf{Q2: What are the linear coefficients?} \label{section::coeff}
We  measure the decay rate  $r(\bs{k})=-\df{\ln |\uk|}{t}$ to probe the PDE coefficients. For (\ref{uk::firstorder}), $r(\bs{k})= \nu \mk^2-\sigma$. One regression across modes $\bs{k}$ therefore returns both linear coefficients $\nu$ and $\sigma$.  For oscillatory modes, we additionally estimate the frequency $\omega(\bs{k})$ from the spacing of successive minima in $\uk(t)$. For (\ref{uk::2ndorder}), $\omega^2(\bs{k})\approx c^2\mk^2-\sigma$, and the regression can recover the coefficient $c^2$. Since $\omega(\bs{k})$ is only an estimate, we do not use the intercept to determine $\sigma$.

\textbf{Q3: Is the PDE linear or nonlinear?}
We measure two quantities. The first is the fraction of spectral energy above a wavenumber threshold, compared between the start and end of a time window. The second is the coefficient of determination  $R^2$ of the rate regression in Q2. Both quantities can probe the nonlinearity of the PDE. 

A nonlinear term generally couples different modes. For example, the Fourier transform of an advection term $u\pf{u}{x}$ is
\begin{equation}
	\mathcal{F}\left[u\pf{u}{x}\right]_{\bs{k}}= \sum_{\bs{k}_1+\bs{k}_2=\bs{k}}i \bs{k}_{2,x}\hat{u}_{\bs{k}_1}\hat{u}_{\bs{k}_2}, \label{F:adv}
\end{equation}
where $\bs{k}_{2,x}$ is the $x$-component of $\bs{k}_2$. Therefore  mode $\bs{k}$ is driven by products of other modes. Energy consequently migrates toward high wavenumbers, which the first measurement detects. And because the regression in Q2 presumes each mode decays independently, nonlinearity degrades its fit, which the coefficient of determination $R^2$  detects. A low $R^2$ therefore indicates the PDE is likely nonlinear.

\textbf{Q4: Is there advection?}
We measure the gain in spatial correlation under translation. For two time frames  $t_0<t_1$, we compute
\begin{equation}
	g = \max_{\boldsymbol{s}} \left[ \text{corr}\left( u(\boldsymbol{x}, t_0), u(\boldsymbol{x} + \boldsymbol{s}, t_1) \right) - \text{corr}\left( u(\boldsymbol{x}, t_0), u(\boldsymbol{x}, t_1) \right) \right]
	\end{equation}
	together with the maximising shift $\bs{s}^*$. Advection translates structure without deforming it, so a field under transport correlates poorly with itself at zero shift but strongly once shifted by the displacement, giving a large $g$ and a nonzero $\bs{s}^*$. Diffusion and growth act without preferred direction, leaving the maximum at $\bs{s}^*=\bs{0}$ and $g$ near zero. The measurement thus separates directional transport from isotropic evolution, and $\bs{s}^*$ also estimates the advection velocity given the time interval $t_1-t_0$.
	
	\subsection{LLM proposing an equation}
The diagnostics of §\ref{section::dataint} are rendered as text, approximately 300 tokens, and appended to a prompt with two further specifications. First, the symbols an equation may be built from: the field $u$, and its temporal and spatial derivatives. Any algebraic combination is admissible. We deliberately do not include a library of candidate terms. Second, the output form: the LLM reasons in prose, then states its PDE proposal in a delimited block with explicit numerical coefficients. This makes the answer separable from a reasoning thread that typically consists of thousands of tokens.

\subsection{Parsing and canonicalisation}
	The proposed PDE is extracted  and rewritten into a single canonical form, so that expressions differing only in notation score identically. A language model does not adhere to one convention: the same Laplacian may arrive as $d2u/dx2 + d2u/dy2$, $\nabla^2 u$, or $\Delta u$.  Canonicalisation therefore ensures that a correct proposal is scored on its content rather than on its notation.
	
\subsection{Evaluation}

Given a proposed PDE structure, the coefficients are fitted by regularized least squares and the equation is scored by both a weak form residual and a sparsity penalty. This regression only fits coefficients for a proposed PDE structure, and never selects the structure itself. We note that the residual alone would be insufficient, because least squares can assign a near zero coefficient to a redundant term. The sparsity penalty can select the shorter PDE when both explain the data equally well, as a  theorist would.

\section{Experiments}

\subsection{Data and benchmark}
Fields are generated by numerically solving PDEs assembled from a library of eight terms. For a scalar field  $u(\bs{x},t)$ on a periodic domain, a first-order sample takes the form $\pf{u}{t}=\nu \nabla^2 u +\bs{a}\cdot \bs{\nabla}u+\sigma u + \beta u\pf{u}{x}+r(u-u^2)$, and a second-order example $\pft{u}{t}=c^2\nabla^2 u + \sigma u +\gamma \pf{u}{t}$. During  data generation, each spatial term is included or omitted at random and its coefficient drawn from a fixed range. These terms are the standard building blocks of transport and kinetics in the molecular sciences. Their combinations recover the reaction-diffusion systems underlying pattern formation, the convection-diffusion equations of  flow dynamics, and the damped wave equations of acoustic and elastic responses. Each field is integrated on a 64 by 64 grid over 50 frames.

\begin{table}[t]
	\centering
	\caption{\textbf{Benchmark strata and accuracy results.} Strata vary along two
		dimensions, the temporal order and the nonlinear terms within. \emph{Self advection, suppressed} and \emph{wave, weakly identifiable}
		are deliberately unfavourable controls where the data  does not determine the
		answer.}
	\label{tab:strata}
	\begin{tabular}{llccc}
		\toprule
		Stratum & Temporal order & Nonlinear term & Number of samples & $F_1$ \\
		\midrule
		Linear only                  & first  & ---                     & 6  & 1.000 \\
		Reaction                     & first  & $r(u-u^2)$              & 4  & 0.500 \\
		Self advection, observable   & first  & $\beta u\,\partial_x u$ & 8  & 0.579 \\
		Self advection, suppressed   & first  & $\beta u\,\partial_x u$ & 4  & 0.625 \\
		Wave                         & second & ---                     & 11 & 0.818 \\
		Wave, weakly identifiable    & second & ---                     & 11 & 0.685 \\
		\bottomrule
	\end{tabular}
\end{table}

Accuracy is quantified by the overlap between the proposed and true term sets. Denoting 
$T$ as the terms of the generating equation and  $P$ for those of the proposal, and $|\cdot|$ for the number of elements in a set,  the accuracy is measured by $F_1=2\frac{|T \cap P|}{|T|+|P|}$.
We report two metrics. Mean $F_1$ measures how much of each equation was identified. Exact recovery is the fraction of exact-matching samples with  $P=T$.  We report both because at this sample size exact recovery is coarse, while $F_1$ can be inflated by a guess containing whichever terms are common. The fixed-term set floor quantifies that inflation.

Samples are drawn in fixed proportions across six strata. The strata vary along two dimensions, the temporal order of the PDE and which nonlinear terms it contains, and are listed with the accuracy in Table \ref{tab:strata}. Two are deliberately unfavourable. In ``self advection, suppressed", the field decays exponentially, so a term scaling as $u^2$ in (\ref{F:adv}) is negligible for most of the observation window whether or not it appears in the equation. In ``wave,  weakly identifiable",  the PDE contains two terms that are difficult to distinguish in data. 

\subsection{Results and discussion}

We compare three inputs to the LLM, with results given in Table \ref{tab:conditions}. The first is the data interpretation of section \ref{section::dataint}, approximately 310 tokens for each sample. The second is a permuted interpretation, computed from a different sample, which controls against the model inferring a plausible equation without consulting its input. The third replaces the interpretation with ten 1D slices of the raw field, approximately 1,230 tokens. As a floor we also report the best fixed term set: the single equation achieving the highest mean accuracy across all samples, which uses no information from any individual field and requires no model call. Table \ref{tab:conditions} shows that data interpretation exceeds the floor by 0.329 at paired Wilcoxon $p=2.6\times 10^{-5}$ and doubles exact recovery at $p=6.6\times 10^{-3}$.

\begin{table}[t]
	\centering
	\caption{\textbf{Accuracy by inputs.} All results use the same benchmark, model,
		token budget and parser, and differ only in what the model is shown about the
		field. The floor is the single fixed term set achieving the highest mean $F_1$
		across all samples.}
	\label{tab:conditions}
	\begin{tabular}{lccc}
		\toprule
		Input & $F_1$ & Exact & Unscoreable \\
		\midrule
		\textbf{Data interpretation}   & \textbf{0.720} & \textbf{14/44} & 0/44 \\
		Best fixed term set (floor)    & 0.391          & 7/44           & --- \\
		Permuted interpretation        & 0.244          & 4/44           & 0/44 \\
		Raw field slices               & 0.225          & 2/43           & 15/44 \\
		\bottomrule
	\end{tabular}
\end{table}

Interpreting the field is substantially more effective than showing the raw data. Data interpretation reaches $F_1=0.720$ against 0.225 for raw field slices, with exact recovery on 14 of 44 samples against 2. The gain in accuracy comes from the interpretations. Permuting the assignment between interpretations and fields, so that each sample receives another sample's measurements, reduces accuracy from 0.720 to 0.244 ($p<3\times 10^{-7}$). This collapse shows the LLM reads the interpretations and follows them, even to a wrong answer. In Table \ref{tab:conditions}, both controls fall below the optimized fixed equation, which is consistent with the LLM not relying on common terms when its input is uninformative.

Accuracy is highest where the interpretation is most complete. In Table \ref{tab:strata}, the ``linear only" stratum is recovered perfectly, consistent with the dispersion fit  in Q2. The shortfall concentrates on the nonlinear strata, where the interpretation establishes that a nonlinear term is present but cannot determine its form.

\section{Concluding remarks}

\textbf{Limits}. The limits of this work are mainly on  the small scale of datasets and on the insufficient characterization tools in data interpretation. On scale, this is a workshop paper: 44 samples, one frozen model, and noise free simulated fields drawn from a library of eight terms. That being said, the differences we report are large, and the paired tests are exact rather than asymptotic, so the sample size is adequate to establish them. On characterization, the data interpretation defines the space of PDE structures that can be discovered. Our characterizations establish that a nonlinear term is present without indicating its form, and the model consequently identifies those terms less accurately. A richer set of characterization tools would expand the space of discoverable PDEs; the ones reported here were chosen to test the idea rather than to exhaust it. The characterizations are also fragile to noise, so robust differentiation and noise filtering are required before being applied to experimental data. 

\textbf{Conclusions}. Our results show that an LLM given a small set of physical measurements of a field identifies its governing equation substantially better than one  given the field itself.  The measurements cost a fraction of a second per field, use fewer tokens than the raw data they replace, and require no training. Therefore, data interpretation is a compact and useful input for LLM-driven symbolic regression. More broadly, this pipeline is promising to automate the top-down route to field theory construction. Identifying which mechanisms underlie field data, and quantifying whether a candidate equation accounts for it, are the two key steps that conventionally take most time and effort for a theorist. In our LLM-driven symbolic regression, the data interpretation performs the first and the evaluator the second. Because neither is trained or tailored to a particular system, a new field can be analyzed in minutes rather than months, and every proposal is tested against the data as it is made. Overall, this pipeline opens a path to theory construction that scales with experiment: as a laboratory varies molecular components across thousands of systems, each new dataset could automatically yield  a candidate theory, already validated against the data that produced it.

\textbf{Future work}. A direct extension of our work is to apply data interpretation to the residual, as shown in dotted lines in Fig. \ref{fig1}.  Given a proposed equation, the part of the dynamics it fails to account for is itself a field, and interpreting that field would indicate what the proposal is missing. The LLM could then revise its answer, and the process could iterate until the equation residual is minimal.  Three further directions follow. Robust differentiation would test whether the interpretation survives experimental noise, a precondition for any use on experimental data. Prepending the data interpretation to an existing evolutionary search would test whether it improves such frameworks without modification to the search itself. And a richer set of characterization tools would widen the space of PDEs that can be reached.

\bibliographystyle{unsrt}
\bibliography{topdown}

\end{document}